\documentclass{article}

\usepackage{microtype}
\usepackage{booktabs}
\usepackage{caption}

\usepackage{graphicx}
\usepackage{subfigure}
\usepackage{multirow}
\usepackage{rotating}
\usepackage{pgfplots}
\pgfplotsset{compat=1.17}
\usetikzlibrary{arrows.meta}

\usepackage{hyperref}

\usepackage[accepted]{mlsys2025}

\makeatletter
\renewcommand{\printAffiliationsAndNotice}[1]{%
\stepcounter{@affiliationcounter}%
{\let\thefootnote\relax\footnotetext{\hspace*{-\footnotesep}#1%
\forloop{@affilnum}{1}{\value{@affilnum} < \value{@affiliationcounter}}{
\textsuperscript{\arabic{@affilnum}}\csname @affilname\the@affilnum\endcsname}.
Correspondence to: \mlsyscorrespondingauthor@text.
}}}
\makeatother
\hypersetup{pdfsubject={}}

\mlsystitlerunning{DWISE: Denoising Workload-Informed Scheduling Engine}

\begin{document}

\twocolumn[
\mlsystitle{DWISE: Denoising Workload-Informed Scheduling Engine}

\begin{mlsysauthorlist}
\mlsysauthor{Lie Li}{bris}
\mlsysauthor{Wen Li}{bris}
\mlsysauthor{Junxiao Shen}{bris}
\mlsysauthor{Jindong Gu}{ox}
\mlsysauthor{Guosheng Hu}{bris}
\end{mlsysauthorlist}

\mlsysaffiliation{bris}{University of Bristol, Bristol, United Kingdom}
\mlsysaffiliation{ox}{University of Oxford, Oxford, United Kingdom}

\mlsyscorrespondingauthor{Guosheng Hu}{huguosheng100@gmail.com}

\mlsyskeywords{Diffusion Language Models, LLM Serving, Request Scheduling, Workload Prediction, Mixed-Phase Batching}

\vskip 0.3in

\begin{abstract}
Unlike autoregressive  large language models (AR-LLMs), diffusion LLMs (dLLMs) generate text by iteratively denoising \emph{many} tokens in parallel, offering a promising path to efficient inference. Existing serving frameworks, however, schedule dLLM requests with policies designed for AR models, namely first-come-first-served (FCFS) admission and prefill priority. We find that both policies are ill-suited to dLLMs, wasting much of their speed. FCFS ignores that a dLLM request's decoding cost is highly heterogeneous and unknown at arrival, causing severe head-of-line blocking. Prefill priority, widely used in AR-LLM serving to keep time-to-first-token low, sharply cuts dLLM efficiency, as rounds that run only prompts leave most of the batch idle. We present DWISE, a Denoising Workload-Informed Scheduling Engine for efficient dLLM serving. DWISE schedules requests by their remaining work, measured in a block-level unit shared by prefill and denoising, so both phases share a  round. To predict the unknown denoising workload, DWISE combines a lightweight prompt-based predictor at admission with a single refinement from the first denoising round's hidden states, which capture model-specific information absent from the prompt and require no additional model execution. Implemented in SGLang and evaluated on five models from three dLLM families under bursty workloads, DWISE reduces mean end-to-end latency over SGLang by 5.4$\times$ on average and up to 14.9$\times$ while preserving task accuracy. The code will be released publicly.

\end{abstract}

]

\printAffiliationsAndNotice{}

\section{Introduction}
\label{sec:introduction}

Recently, diffusion large language models (dLLMs) have emerged rapidly as an alternative to autoregressive (AR) LLMs~\citep{NEURIPS2020_1457c0d6}. Since 2025, they have grown from 8B research models~\citep{nie2025largelanguagediffusionmodels} to 100B-parameter releases~\citep{bie2025llada20scalingdiffusionlanguage, bie2026llada21speedingtextdiffusion}, and Google has released the open-weight DiffusionGemma, fine-tuned from Gemma~4~\citep{diffusiongemmateam2026diffusiongemmatechnicalreport}. Instead of emitting one token at a time, a dLLM iteratively denoises a fully masked sequence, resolving many tokens in parallel per round~\citep{NEURIPS2021_958c5305, sahoo2024mdlm, shi2024md4}. At comparable model scale, dLLMs are now competitive with AR models on language, mathematical reasoning, and coding tasks~\citep{ye2025dream7bdiffusionlarge}, while parallel decoding removes the token-by-token bottleneck for much faster generation~\citep{labs2025mercuryultrafastlanguagemodels}.

This speed matters most in deployment, where a model serves large and variable streams of requests, making inference efficiency a major systems concern: already in 2021, inference took 70\% of Meta's AI infrastructure power capacity, 3.5$\times$ that of training~\citep{wu2022sustainableaienvironmentalimplications}. dLLMs thus offer a promising path to more efficient inference than AR LLMs, and the serving system must turn faster generation into high throughput and low latency. Serving systems have begun to adapt: frameworks such as SGLang~\citep{zheng2024sglangefficientexecutionstructured} now support diffusion decoding, and dedicated dLLM systems target memory~\citep{fan2026dllmserve}, recurring prefills from KV refresh~\citep{sangam2026}, and SLO attainment~\citep{chang2026tediserve}. Yet request scheduling in these frameworks has not adapted: under policies inherited from AR serving, prefill takes priority over decoding, in rounds of its own, leaving most of the batch empty while running requests wait. Long-context agentic workloads~\citep{bie2026llada22} make this worse: longer prompts take more such rounds. In serving, dLLMs thus lose much of their speed, and recovering it requires scheduling built for how they generate text, which remains largely unexplored.

These inherited policies decide which requests run, wait, or retain accelerator resources per iteration~\citep{orcayu2022}, and thus affect serving throughput and request latency. Two of them make requests wait: first-come-first-served (FCFS) admission~\citep{quenuesystem}, which ignores execution cost, and prefill priority~\citep{agrawal2024tamingthroughputlatencytradeoffllm}, under which processing new prompts (prefill) precedes generating output (decoding). FCFS is attractive because it requires no prior knowledge of request cost, but under heterogeneous workloads it allows long-running requests to occupy scarce batch capacity while shorter requests queue behind them, and the resulting head-of-line (HOL) blocking inflates latency~\citep{wu2024fastdistributedinferenceserving}. Multi-level feedback queues (MLFQ) mitigate this blocking without such knowledge by demoting requests as they run, so short requests finish first~\citep{MLFQ}. The wait under prefill priority, in contrast, is a deliberate trade-off that pays off for AR models: running requests pause for one step, but the new request gets its first token in that same step.

In practice, a dLLM denoises its output one block at a time, not all at once~\citep{arriola2025block}. A block is a fixed-size span of consecutive tokens (e.g., 32). Each output block starts fully masked, and every denoising round unmasks, in parallel, the tokens the model is confident about~\citep{wu2025fastdllmtrainingfreeaccelerationdiffusion}, so a block needs a variable number of rounds. During serving, prompts are also split into blocks, and each round is one batched forward pass over one block per selected request: a prompt block (prefill) or its current output block (decoding). Hence, dLLMs break two properties AR scheduling relies on: that output length measures work, and that processing a prompt yields the first token.

\begin{figure}[t]
    \centering
    \includegraphics[width=\columnwidth]{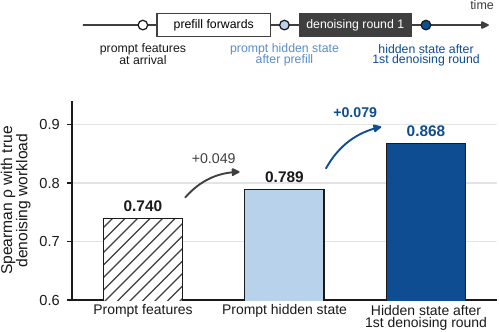}
    \caption{\textbf{The first denoising round reveals remaining work.} Spearman $\rho$~\citep{man1904proof} between true denoising workload and estimates from prompt features, the prompt hidden state, and the first-round hidden state on LLaDA2.1-mini (five-fold CV).}
    \label{fig:work_signal}
\end{figure}

Losing these properties raises two scheduling challenges: (1) Accurate workload prediction, which efficient scheduling requires, is hard for dLLMs. Prefill work is determined by the prompt length, but decoding time depends on the number of denoising rounds, which is unknown at arrival and highly heterogeneous, ranging from a few to hundreds of rounds across requests. Length-based LLM schedulers instead predict output length from the prompt~\citep{jin2023s3increasinggpuutilization,zheng2023responselengthperceptionsequence,fu2024efficientllmschedulinglearning} or during generation~\citep{shahout2024trail}. A size-based scheduler therefore has no reliable estimate to act on. (2) For dLLMs, prefill priority keeps the cost but loses the benefit: the first output block still needs several rounds, which later prompts keep deferring. Admitting a new request stalls running requests for as many rounds as its prompt has blocks. Yet prompt and output blocks can share a round, so both challenges concern one decision: which blocks run each round.

To address these challenges, we present DWISE, a Denoising Workload-Informed Scheduling Engine that predicts denoising workload from the model's own hidden states and schedules prefill and decoding as the same block-level work. For challenge (1), we observe that a prompt-based predictor sees only the input tokens and misses how the model will process them. The model's hidden states capture both, and those after the first denoising round predict the remaining rounds significantly better than any prompt-side signal (Figure~\ref{fig:work_signal}). Since every request runs this round anyway, DWISE estimates denoising workload from prompt features at admission and refines the estimate once from the first-round hidden states, without extra model execution. For challenge (2), DWISE drops prefill priority: it measures a request's remaining work in blocks, as its remaining prefill blocks plus its estimated denoising rounds, and fills each round with the requests that have the least remaining work, whether they are in prefill or decoding. New arrivals thus no longer stall running requests, and short requests finish first.

In summary, we make the following contributions.
\begin{itemize}

\item Based on the intrinsic difference between block-wise (dLLM) and token-wise (AR) text generation, we design DWISE, to our knowledge the first dLLM scheduler that orders requests by their remaining work measured in denoising rounds, not output tokens. Instead of prioritizing prefill as AR serving does, DWISE counts prefill blocks in the same unit and serves the requests with the least remaining work first.

\item We show that a dLLM's first denoising round reveals what the prompt cannot. Its hidden state raises the rank correlation with remaining work from 0.74 to 0.87 (Figure~\ref{fig:work_signal}), and from 0.16 to 0.64 within a request category, where prompt features barely order requests (Table~\ref{tab:loco}). DWISE refines each estimate once, and the gain holds for unseen categories.

\item We implement \textsc{DWISE} in SGLang and evaluate it on five models from three dLLM families against FCFS and five prior LLM schedulers. It achieves the lowest mean latency in every setting, reducing it over SGLang by $5.4\times$ on average (up to $14.9\times$) and raising throughput by up to $2.2\times$, with accuracy within one point.

\end{itemize}

\section{Background and Related Work}

\subsection{dLLM Serving}

\textbf{Diffusion Language Models.}
Diffusion models \citep{ho2020ddenoising,sohldickstein2015deepunsupervisedlearningusing}, first popularized in computer vision, generate through iterative denoising, and discrete diffusion formulations \citep{NEURIPS2021_958c5305} extend them to language tokens. Recent dLLMs, such as the LLaDA family \citep{nie2025largelanguagediffusionmodels}, scale masked diffusion to large language models and denoise multiple token positions in parallel. To retain this parallelism while supporting causal, variable-length generation, block-diffusion models \citep{arriola2025block} factorize the output into a sequence of blocks: blocks are generated causally, while tokens within each block are denoised in parallel.

\textbf{Serving Execution.} To sustain high GPU utilization under dynamic request arrivals, LLM serving systems use continuous batching, as in vLLM \citep{kwon2023efficient} and SGLang \citep{zheng2024sglangefficientexecutionstructured}. In autoregressive models, requests first process their prompts during prefill and then generate one token per decoding iteration. Prefill and decoding differ in execution characteristics \citep{patel2024splitwiseefficientgenerativellm,zhong2024distservedisaggregatingprefilldecoding}, motivating chunked prefill in Sarathi-Serve \citep{agrawal2024tamingthroughputlatencytradeoffllm} to reduce interference. dLLMs, however, generate through iterative denoising \citep{wu2025fastdllmtrainingfreeaccelerationdiffusion,liu2026dllmcacheacceleratingdiffusionlarge}. In SGLang's dLLM path, each serving round is one batched forward pass, and each selected request processes one block, prompt or output. Prefill advances through a known number of blocks, whereas decoding an output block takes a variable number of denoising rounds. We refer to the rounds a request still needs as its denoising workload.

\subsection{Scheduling Optimization for LLM Serving}
\label{scheduling_optimization}

\textbf{Request Scheduling in Shared Systems.}
When job sizes are known, shortest-job-first (SJF) \citep{SJF} prioritizes shorter jobs, while shortest-remaining-processing-time (SRPT) \citep{SRPT} extends this principle through preemption and minimizes mean response time under standard assumptions. In practice, however, exact job sizes are often unavailable before execution. Non-clairvoyant policies such as multilevel feedback queues (MLFQ) \citep{MLFQ} and other age-based policies \citep{scully2018soapcleananalysisagebased} infer job characteristics from observed execution progress, while prediction-based policies \citep{mitzenmacher2019schedulingpredictionspricemisprediction} replace unknown service times with estimated ones.

\textbf{Request Scheduling for LLM Serving.}
LLM serving faces the same problem, since generation cost is unknown at arrival. FastServe~\citep{wu2024fastdistributedinferenceserving} infers priority from attained service with a preemptive skip-join MLFQ. Prediction-based approaches estimate size from the prompt: S$^{3}$~\citep{jin2023s3increasinggpuutilization} and response-length-aware scheduling~\citep{zheng2023responselengthperceptionsequence} predict output length, and LTR~\citep{fu2024efficientllmschedulinglearning} predicts relative order. TRAIL~\citep{shahout2024trail} refines the estimate during execution from intermediate representations. All of them use remaining output tokens as a proxy for remaining work. This proxy does not hold for dLLMs, whose output blocks each take a variable number of denoising rounds. dLLM serving systems have so far targeted memory~\citep{fan2026dllmserve}, recurring prefills from KV refresh~\citep{sangam2026}, or SLO attainment~\citep{chang2026tediserve}, not ordering by remaining work.

\section{Motivation}
\label{motivation}

For a dLLM request, the work that matters to scheduling is not how many tokens it will produce but how many denoising rounds it still needs. Policies inherited from AR serving overlook three properties of this workload: first, it varies by more than 100$\times$ across requests (Section~\ref{sec:motiv-hetero}); second, the first denoising round already predicts it far better than the prompt does (Section~\ref{sec:motiv-signal}); and third, its blocks occupy the same kind of slot in a round as prefill blocks (Section~\ref{sec:motiv-unified}).

\subsection{Denoising Workload Is Highly Heterogeneous}
\label{sec:motiv-hetero}

\begin{figure}[t]
    \centering
    \includegraphics[width=\columnwidth]{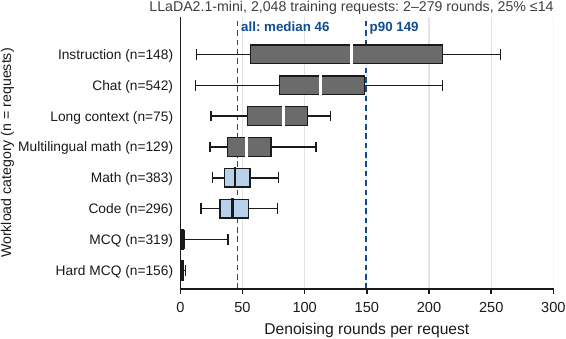}
    \caption{Denoising workload varies substantially across requests (LLaDA2.1-mini). Boxes p25--p75, line median, whiskers p5--p95.}
    \label{fig:work_distribution}
\end{figure}

Denoising workload varies by up to two orders of magnitude across requests in the models and workloads we study. In the LLaDA2.1-mini workload of Figure~\ref{fig:work_distribution}, a quarter of requests finish within 14 denoising rounds, the median needs 46, and the 90th percentile needs 149. Task type explains part of this spread: instruction-following and chat requests dominate the long tail, yet requests of the same task type still span a wide range of rounds. This heterogeneity is the condition under which size-based scheduling pays off: under arrival-order scheduling, a request that needs 14 rounds can wait behind one that needs 149. Exploiting it requires the right measure of size. Two requests that emit the same number of blocks can consume very different amounts of GPU time, because each block is resolved over its own number of denoising rounds. Generation-side work in dLLMs should therefore be measured in denoising rounds rather than output tokens.

\subsection{First Denoising Round Reveals Remaining Work}
\label{sec:motiv-signal}

A request's remaining denoising workload is unknown at arrival, but one round of execution reveals much of it. At admission, a scheduler sees only the prompt, which is what existing length-aware schedulers predict from. The prompt, however, carries only the input tokens, not how the model will process them. Under the threshold-based parallel decoding used by LLaDA2.0~\citep{bie2025llada20scalingdiffusionlanguage}, a block finishes once every position's predicted token clears the confidence threshold, so the number of rounds it takes depends on how quickly the model's confidence rises during denoising, a trajectory that becomes visible only once denoising begins.

The first denoising round exposes this trajectory. On LLaDA2.1-mini (Figure~\ref{fig:work_signal}), the hidden state after this round reaches a Spearman correlation of 0.868 with true denoising workload, against 0.740 for prompt features and 0.789 for the prompt hidden state, and the same holds across all five models we study (Appendix~\ref{app:models}). The signal is also free: the first round is a forward pass the request executes anyway. A dLLM scheduler can therefore rank a request from its prompt at admission and correct that ranking once, after the first denoising round. This correction comes early enough to matter. After one round, the median request in Figure~\ref{fig:work_distribution} still has 45 of its 46 rounds ahead, so a request whose prompt-based estimate was wrong can still be reordered before most of its work is done.

\begin{figure}[t]
    \centering
    \includegraphics[width=\columnwidth]{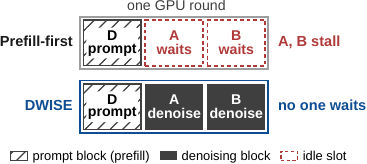}
    \caption{Prefill-first execution can stall denoising, while both phases share the same block-level scheduling granularity.}
    \label{fig:joint_round}
\end{figure}

\subsection{Prefill and Denoising Blocks Can Share a Round}
\label{sec:motiv-unified}

A request's remaining service also includes prompt prefill, which, unlike denoising, is known exactly from the number of unprocessed prompt blocks. SGLang's dLLM path inherits the prefill-first policy of autoregressive serving, where admitting new prompts quickly protects first-token latency~\citep{gao2024costefficientlargelanguagemodel,298685}. Each serving round is either a prefill round or a denoising round, and whenever prompts are waiting, the round goes to prefill and advances no denoising requests. Under a backlog, prompts keep coming, so denoising is interrupted round after round. A short request that remaining-work ordering would finish first instead waits behind new prompts. Prefill rounds also waste capacity: a waiting prompt enters the pool only as a running request finishes, and each prompt advances one block per round, so a prefill round carries one or two blocks and leaves its other slots idle (Figure~\ref{fig:joint_round}). Section~\ref{sec:eval-breakdown} shows that these nearly empty rounds dominate SGLang's runtime.

Block diffusion removes the reason for this separation. AR serving separates the two phases because they differ in cost: a prefill step processes a whole prompt, a decode step only one token, so mixing them slows decoding. In block diffusion, a prefill step and a denoising step each process one block of tokens against the KV cache of earlier blocks, so they cost the same and can share a round. The idle slots of a prefill round can then carry denoising blocks, and a scheduler can fill each round by how much work each request has left, whichever phase it is in.

\begin{figure}[t]
  \centering
  \includegraphics[width=\columnwidth]{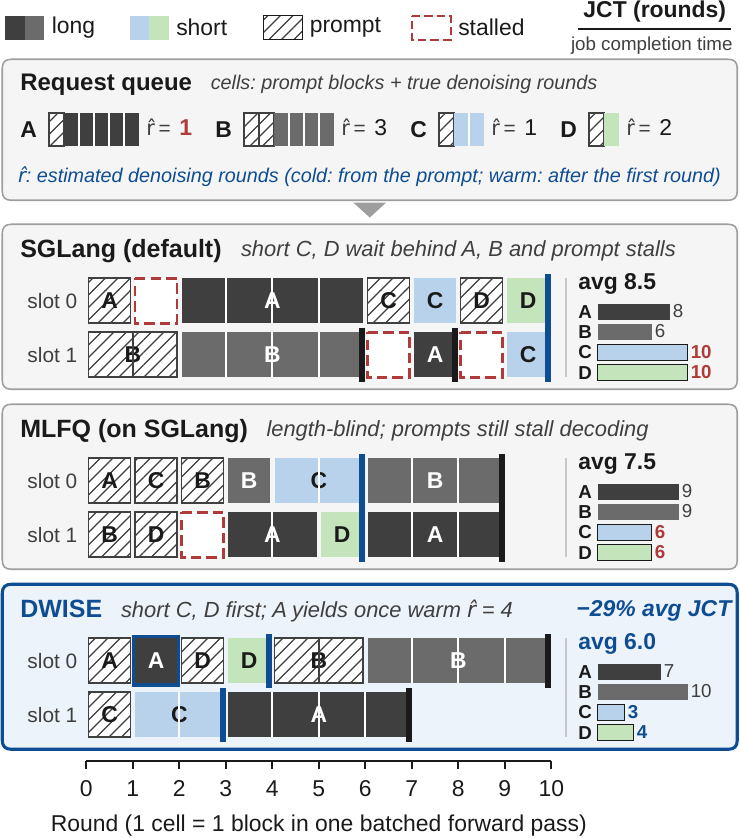}
  \caption{Overview of \textsc{DWISE}.
    Toy example illustrating block-level scheduling under SGLang, MLFQ, and \textsc{DWISE}.}
  \label{fig:intro}
\end{figure}

\section{DWISE: Key Ideas}

\begin{figure}[t]
    \centering
    \includegraphics[width=\columnwidth]{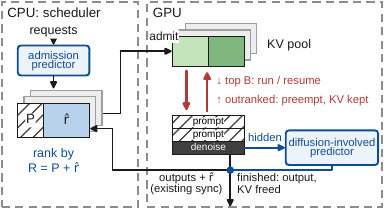}
    \caption{DWISE per-round scheduling loop.}
    \label{fig:loop}
\end{figure}

DWISE is a runtime scheduler for block-diffusion LLM serving. Its key idea is to treat a prefill block and a denoising block as the same unit of work, since the two cost the same. A request's remaining work is its unprocessed prefill blocks plus its remaining denoising rounds, and both phases compete for the same block slots of each forward pass. Since denoising rounds are unknown at arrival, DWISE first estimates them from the prompt. Before each round, it ranks requests by remaining work and fills the slots from the top, pausing any request that drops out, with its KV cache kept. After the first denoising round, a diffusion-involved predictor, a small network on the model's final layer trained offline in under three minutes, refines this estimate from the hidden states. Figure~\ref{fig:intro} shows this loop on four requests: SGLang and MLFQ leave short requests C and D behind longer requests and prompt rounds, whereas DWISE finishes them first and lowers average completion time from 8.5 rounds to 6.0. Since DWISE only decides which blocks enter each round, it composes with decoding-level accelerations such as KV caching and parallel decoding~\citep{wu2025fastdllmtrainingfreeaccelerationdiffusion}; all evaluated models already run with both. Figure~\ref{fig:loop} shows how the loop splits across CPU and GPU.

\subsection{Round-Based Workload Accounting}
\label{sec:ki-unit}

DWISE expresses request demand in the unit the runtime schedules. A remaining prefill block and a remaining denoising round each occupy one slot of one future round, so DWISE counts them equally. The remaining work of request $i$ is

\begin{equation}
R_i = P_i + \hat{r}_i,
\label{eq:remaining-work}
\end{equation}

where $P_i$ is the exact number of unprocessed prefill blocks and $\hat{r}_i$ is the estimated number of remaining denoising rounds. Size-based policies such as SRPT presuppose that all jobs' remaining service is measured on one scale~\citep{harchol2013performance}. Eq.~(\ref{eq:remaining-work}) supplies that scale, so DWISE can compare a request still in prefill with one already denoising. Which term dominates depends on the block size, so both are needed.

\subsection{In-Path Workload Estimation}
\label{sec:ki-est}

DWISE estimates $\hat{r}_i$ twice, once from each signal in Section~\ref{sec:motiv-signal}. At admission, the admission predictor on the CPU estimates it from prompt features.
After the first denoising round, the diffusion-involved predictor refines it from the final-layer hidden states. It runs asynchronously on the GPU, and its output is collected at the existing end-of-round synchronization, so refinement adds neither model execution nor a synchronization point.
One refinement suffices: the first round already exposes how the request will denoise, and refining more often only adds preemptions (Section~\ref{sec:eval-breakdown}).

\textbf{Prompt-based Predictor.} Our admission predictor is a LightGBM regressor~\citep{lightgbm} (100 trees, at most 15 leaves) trained over three prompt features: word count, newline count, and token count after the chat template. It predicts total denoising rounds in log space, and its inference takes 29~$\mu$s per request on the CPU. Among the regressors we compared, it ranks best at the lowest cost (Appendix~\ref{app:estimator}).

\textbf{Diffusion-involved Predictor.} The diffusion-involved predictor
takes the final-layer hidden rows of the first denoising round, the model's parallel draft of the whole block, mean-pooled into one vector and concatenated with the normalized prompt length and output budget. It is a discrete-time survival model~\citep{gensheimer2019scalable}: a two-layer MLP (width 256, GELU, dropout 0.2) that outputs one hazard logit per future round up to $K$, the largest round count in training ($K{=}279$ on LLaDA2.1-mini), and the refined estimate is the mean of the resulting distribution over remaining rounds. An isotonic mapping~\citep{barlow1972isotonic}, fitted on the training split, maps this output to actual remaining rounds, the scale the admission predictor targets, so requests holding admission and refined estimates are ranked consistently in Eq.~(\ref{eq:remaining-work}). The predictor averages three such MLPs trained with different seeds.
Together they total 1.79M parameters (0.01\% of the model) and take about 0.8\,ms of GPU time per round, under 1\% of the 95~ms median round, so refinement is effectively free.

\subsection{Preemptive Mixed-Phase Batching}
\label{sec:ki-batch}

Each serving round runs one batched forward pass over $B$ block slots. DWISE selects the $B$ resident requests with the smallest $R_i$ and packs the next block of each into the batch, prompt or denoising. Prefill and denoising thus share every forward pass, and new prompts no longer stall generation. DWISE recomputes the ranking every round, so a running request is displaced as soon as it is outranked, either by a newly arrived shorter request or by a refinement that raises its own estimate. A displaced request is paused with its KV cache and partially denoised block kept on the GPU rather than swapped out~\citep{298685}, and resumes without recomputation. The policy is preemptive SRPT~\citep{SRPT} with the serving round as its quantum.

Preemption pays off only with mixed-phase batching: under prefill-first, each prefill round stalls all residents. It is safe because residency is controlled separately from execution. A waiting request becomes resident only when it ranks among the top $B$ and the worst-case KV footprint of all residents still fits in the pool, with at most $kB$ residents for a resident-set multiplier $k$.
Every resident request is thus guaranteed the memory to finish, so preemption never overcommits or deadlocks the pool.

\section{System Implementation}
\label{sec:system}

We implement DWISE in SGLang's dLLM serving path as a runtime patch, without modifying SGLang's source. The implementation splits along SGLang's existing boundary between scheduling and execution and has two parts: a CPU-side scheduler that takes over admission and batch formation, and a GPU-side refinement path that evaluates the diffusion-involved predictor inside the forward pass. All other components, such as the model, denoising kernels, KV memory manager, and CUDA-graph path, stay unchanged.

\begin{algorithm}[t]
\caption{DWISE per-round scheduling.}
\label{alg:dwise}
\begin{algorithmic}[1]
\REQUIRE waiting set $\mathcal{W}$, resident set $\mathcal{S}$, block slots $B$, resident-set multiplier $k$
\FOR{each serving round}
  \STATE sort $\mathcal{W} \cup \mathcal{S}$ by $R_i = P_i + \hat{r}_i$
  \FOR{$i \in \mathcal{W}$ in sorted order}
    \IF{$i$ ranks in the top $B$ \AND $|\mathcal{S}| < kB$ \AND its KV reservation fits}
      \STATE move $i$ from $\mathcal{W}$ to $\mathcal{S}$
    \ENDIF
  \ENDFOR
  \STATE $\mathcal{E} \gets$ $B$ requests in $\mathcal{S}$ with the smallest $R_i$
  \STATE pause $\mathcal{S} \setminus \mathcal{E}$
  \STATE run one forward pass over the next blocks of $\mathcal{E}$
  \FOR{$i \in \mathcal{E}$}
    \IF{$i$ processed a prompt block}
      \STATE $P_i \gets P_i - 1$
    \ELSIF{$i$ finished its first denoising round}
      \STATE $\hat{r}_i \gets$ refined estimate
    \ELSE
      \STATE $\hat{r}_i \gets \max(\hat{r}_i - 1,\, 0)$
    \ENDIF
  \ENDFOR
  \STATE drop finished requests from $\mathcal{S}$ and free their KV
\ENDFOR
\end{algorithmic}
\end{algorithm}

\subsection{Scheduler Integration}
\label{sec:impl-sched}

The scheduler keeps three fields per request: the unprocessed prompt blocks $P_i$, the remaining-rounds estimate $\hat{r}_i$, and a residency state (waiting, running, or paused). Algorithm~\ref{alg:dwise} shows the resulting per-round loop. DWISE intervenes at four points of each serving round. At admission, it applies the global remaining-work order together with the KV reservation policy of Section~\ref{sec:ki-batch}. At batch formation, it picks the next block of the highest-priority resident requests, which removes SGLang's prefill-first phase barrier. At result processing, SGLang already walks the batch to commit outputs, and DWISE updates the three fields in that same pass. After the first denoising round, it records each request's refined estimate once.

None of these steps touches memory management. The token pool, the radix prefix cache, and the pause/resume path are reused, so reordering execution needs neither a separate memory manager nor reconstruction of request state. Keeping the control loop on the CPU is deliberate. Its per-round work is many small, irregular operations, and an earlier GPU-resident planner was slower because kernel-launch cost dominated at that granularity.

\subsection{GPU Refinement Path}
\label{sec:impl-gpu}

The diffusion-involved predictor is the one component that runs on the GPU, since its input is the hidden state of the first denoising round. The round's final-layer hidden rows are pooled with tensor operations and passed through the predictor in one batched GEMM for every request refined in that round. The predictor's scalar inputs, normalized prompt length and output budget, arrive through pinned host memory without blocking, and the predictor is enqueued on the forward pass's stream. Because the hidden-state operations and refinement hooks are registered before CUDA-graph capture, SGLang's graph-replay path stays enabled.

\section{Evaluation}
\label{evaluation}
\subsection{Experiment Settings}
\label{sec:eval-setup}

\textbf{Hardware.} Each serving instance runs on one NVIDIA GH200 Grace Hopper Superchip, which pairs a Hopper GPU (132 SMs, 96\,GB of HBM3) with a 72-core Arm Neoverse-V2 Grace CPU over NVLink-C2C.

\textbf{Software and configuration.} All policies run on SGLang v0.5.20's dLLM path. Within each model, policies use the same block slots per round ($B=128$; $B=16$ for DiffusionGemma) and the same KV pool (50.8, 65.9, and 13.1\,GiB for LLaDA2.x, SDAR-8B, and DiffusionGemma). DWISE uses $k=8$ (at most $kB$ resident requests), two blocks of reservation slack per resident request, and a 90\% KV reservation limit. We report the mean of three runs.

\begin{table}[t]
\centering
\small
\caption{Serving and predictor-training requests.}
\label{tab:workload}
\setlength{\tabcolsep}{5pt}
\begin{tabular}{@{}lrr@{}}
\toprule
Workload & Serving & Train\\
\midrule
\textit{Open-ended} & 1,973 & 986 \\
\quad ShareGPT~\citep{sharegpt}
    & 1,085 & 542 \\
\quad MBPP~\citep{austin2021programsynthesislargelanguage}
    & 505 & 250 \\
\quad HumanEval~\citep{chen2021evaluating}
    & 87 & 46 \\
\quad IFEval~\citep{zhou2023instructionfollowingevaluationlargelanguage}
    & 296 & 148 \\
\addlinespace[2pt]
\textit{Single-answer} & 1,973 & 987 \\
\quad MATH (Level 5)~\citep{hendrycks2021measuringmathematicalproblemsolving}
    & 421 & 203 \\
\quad GSM8K~\citep{cobbe2021trainingverifierssolvemath}
    & 345 & 180 \\
\quad MGSM~\citep{shi2022languagemodelsmultilingualchainofthought}
    & 258 & 129 \\
\quad MMLU-Pro~\citep{wang2024mmluprorobustchallengingmultitask}
    & 637 & 319 \\
\quad ARC-C~\citep{clark2018thinksolvedquestionanswering}
    & 312 & 156 \\
\addlinespace[2pt]
\textit{Long-context}: Multi-needle~\citep{kamradt2023llmtest_needleinahaystack}
    & 150 & 75 \\
\midrule
\textbf{Total} & \textbf{4,096} & \textbf{2,048} \\
\bottomrule
\end{tabular}
\end{table}

\begin{table*}[t]
\centering
\small
\caption{End-to-end serving performance and task accuracy across models and arrival patterns. B: burst; P: burst+Poisson.}
\label{tab:e2e}

\setlength{\tabcolsep}{3.0pt}
\renewcommand{\arraystretch}{1.05}

\begin{tabular*}{\textwidth}{
@{\extracolsep{\fill}}
l cc cc cc cc cc
@{}
}
\toprule
& \multicolumn{2}{c}{LLaDA2.0-mini}
& \multicolumn{2}{c}{LLaDA2.1-mini}
& \multicolumn{2}{c}{LLaDA2.2-mini}
& \multicolumn{2}{c}{SDAR-8B}
& \multicolumn{2}{c}{DiffusionGemma} \\
\cmidrule(lr){2-3}
\cmidrule(lr){4-5}
\cmidrule(lr){6-7}
\cmidrule(lr){8-9}
\cmidrule(lr){10-11}
& B & P & B & P & B & P & B & P & B & P \\
\midrule

\multicolumn{11}{@{}l}{\textit{Mean latency (s) $\downarrow$}} \\
SGLang
    & 603.1 & 597.4
    & 545.5 & 535.9
    & 576.6 & 571.7
    & 9053  & 8973
    & 325.9 & 262.7 \\
MLFQ
    & 716.0 & 721.8
    & 625.9 & 616.9
    & 651.0 & 654.4
    & 11382 & 12141
    & 346.2 & 282.1 \\
SSJF
    & 739.3 & 728.4
    & 602.7 & 584.6
    & 647.4 & 611.7
    & 11912 & 10936
    & 323.5 & 264.2 \\
LTR
    & 702.7 & 679.6
    & 645.3 & 613.9
    & 664.2 & 597.3
    & 12217 & 11577
    & 326.9 & 266.1 \\
S$^3$
    & 469.2 & 470.3
    & 412.1 & 408.9
    & 417.3 & 410.4
    & 13557 & 12026
    & 321.1 & 257.4 \\
TRAIL
    & 763.3 & 750.3
    & 661.1 & 661.3
    & 679.5 & 683.4
    & 15004 & 14304
    & 328.4 & 264.6 \\
\textbf{DWISE}
    & \textbf{128.5} & \textbf{96.9}
    & \textbf{76.9}  & \textbf{48.5}
    & \textbf{109.4} & \textbf{79.8}
    & \textbf{609}   & \textbf{668}
    & \textbf{292.4} & \textbf{234.8} \\

\addlinespace[3pt]
\midrule

\multicolumn{11}{@{}l}{\textit{Throughput (req/s) $\uparrow$}} \\
SGLang
    & 3.48 & 3.34
    & 3.90 & 3.76
    & 3.65 & 3.49
    & 0.24 & 0.23
    & \textbf{6.22} & \textbf{6.14} \\
MLFQ
    & 3.59 & 3.43
    & 4.13 & 4.00
    & 3.87 & 3.70
    & 0.22 & 0.20
    & 5.88 & 5.85 \\
SSJF
    & 3.31 & 3.26
    & 3.96 & 3.91
    & 3.73 & 3.77
    & 0.22 & 0.22
    & 5.81 & 5.82 \\
LTR
    & 3.45 & 3.44
    & 3.95 & 3.97
    & 3.83 & 4.01
    & 0.22 & 0.22
    & 5.79 & 5.83 \\
S$^3$
    & 3.41 & 3.23
    & 3.86 & 3.67
    & 3.66 & 3.50
    & 0.21 & 0.22
    & 5.83 & 5.92 \\
TRAIL
    & 3.21 & 3.14
    & 3.74 & 3.59
    & 3.47 & 3.33
    & 0.19 & 0.19
    & 6.11 & 6.04 \\
\textbf{DWISE}
    & \textbf{6.39} & \textbf{6.47}
    & \textbf{8.02} & \textbf{8.21}
    & \textbf{6.57} & \textbf{6.65}
    & \textbf{0.25} & \textbf{0.25}
    & 6.07 & 6.03 \\

\addlinespace[3pt]
\midrule

\multicolumn{11}{@{}l}{\textit{Task accuracy (\%) $\uparrow$}} \\
SGLang
    & 45.4 & 45.4
    & 44.2 & 44.1
    & 46.5 & 46.5
    & 55.5 & 55.2
    & 51.3 & 51.3 \\
DWISE
    & 46.2 & 46.2
    & 44.3 & 44.4
    & 46.9 & 46.9
    & 55.4 & 55.4
    & 51.2 & 51.8 \\

\bottomrule
\end{tabular*}
\end{table*}

\textbf{Workloads.} The workload mixes open-ended, single-answer, and long-context requests (Table~\ref{tab:workload}): 4,096 requests for evaluation and a prompt-disjoint set of 2,048 with the same composition, used only to train DWISE and the learned baselines. Long-context requests are synthetic multi-needle retrieval prompts~\citep{kamradt2023llmtest_needleinahaystack, hsieh2024rulerwhatsrealcontext} of about 10K tokens, each embedding 2--4 target facts among distractor sentences and querying all of them. They are prefill-heavy and automatically checkable.

\textbf{Models.} We evaluate five block-diffusion LLMs: three LLaDA2.x-mini models (2.0, 2.1, and 2.2)~\citep{bie2025llada20scalingdiffusionlanguage,bie2026llada21speedingtextdiffusion,bie2026llada22} (16B MoE, 32-token blocks), SDAR-8B-Chat~\citep{cheng2025sdarsynergisticdiffusionautoregressionparadigm} (dense, 4-token blocks), and DiffusionGemma-26B-A4B~\citep{diffusiongemmateam2026diffusiongemmatechnicalreport} (MoE, 256-token blocks, prompt encoded in a separate pass). DWISE runs on all five with the same policy. All models use their official decoding settings, a 256-token output budget, greedy decoding, and bfloat16 precision.

\textbf{Arrival patterns.}
In burst, all 4,096 requests arrive at $t=0$, modeling a backlogged server. In burst+Poisson, half arrive at $t=0$ and the rest follow a Poisson process~\citep{zhang2025jitservesloawarellmserving} with a rate above SGLang's service rate, so new requests keep arriving while the backlog is served. The rates are $\lambda = 16$, 0.85, and 8\,req/s for LLaDA, SDAR, and DiffusionGemma, giving Poisson phases of about 128\,s, 2,410\,s, and 256\,s. Request order and arrival times are generated with fixed seeds, so all policies of a model see the identical arrival trace. Appendix~\ref{app:seeds} shows that DWISE's speedup is stable across seeds.

\textbf{Baselines.} We compare against SGLang's FCFS scheduler with prefill priority~\citep{zheng2024sglangefficientexecutionstructured}, MLFQ following FastServe~\citep{wu2024fastdistributedinferenceserving}, and four learned schedulers retrained on DWISE's 2,048 training requests with labels from FCFS profiling. SSJF~\citep{qiu2024efficientinteractivellmserving}, S$^3$~\citep{jin2023s3increasinggpuutilization}, and LTR~\citep{fu2024efficientllmschedulinglearning} predict output length (LTR: its rank) from the prompt at admission, and TRAIL~\citep{shahout2024trail} refines its remaining-output-length estimate from runtime embeddings every round. All four thus rank requests by output tokens, which do not order dLLM requests: code and math emit more tokens than chat yet need fewer denoising rounds (Appendix~\ref{app:fcfs}). All baselines run on SGLang's prefill-first batch former, as originally designed.

\textbf{Metrics.} Following common LLM serving practice~\citep{kwon2023efficient}, we report mean, p50, and p99 end-to-end request latency and throughput. Because DWISE ranks requests by predicted workload, we evaluate workload prediction by Spearman rank correlation ($\rho$)~\citep{man1904proof} between predicted and true workload. Rank-based LLM scheduling likewise depends on relative request order rather than exact cost~\citep{fu2024efficientllmschedulinglearning}. We also report predictor and scheduler overhead, the predicted-to-actual ratio of remaining work, and task accuracy on single-answer and long-context requests.

\subsection{End-to-End Performance}
\label{sec:eval-e2e}

\textbf{Mean latency and throughput.}
On the LLaDA2.x models (Table~\ref{tab:e2e}), DWISE reduces mean latency over SGLang by 4.7--7.1$\times$ under burst arrivals and 6.2--11.1$\times$ under burst+Poisson arrivals, and raises throughput from 3.34--3.90 to 6.39--8.21 req/s. On SDAR-8B, running long-context requests last lets DWISE reduce mean latency by 13.4--14.9$\times$; the geometric mean over the ten settings is 5.4$\times$. On DiffusionGemma, where the two phases cannot share a round, ordering alone gains 1.1$\times$, against at most 1.02$\times$ for any baseline.
Prior LLM schedulers do not transfer to dLLMs: MLFQ, SSJF, LTR, and TRAIL are slower than SGLang on LLaDA2.x and SDAR-8B (0.60--0.96$\times$), and S$^3$ gains at most 1.4$\times$ (Appendix~\ref{app:fcfs} and Section~\ref{sec:eval-breakdown}). On multiple GPUs (TP=2/4, TP=4+EP=4), DWISE's LLaDA2.1-mini speedup stays at 6.7--12.2$\times$, and with a 2{,}048-token output budget at 4.4--5.2$\times$ (Appendix~\ref{app:multigpu}).

\begin{figure}[t]
    \centering
    \includegraphics[width=\columnwidth]{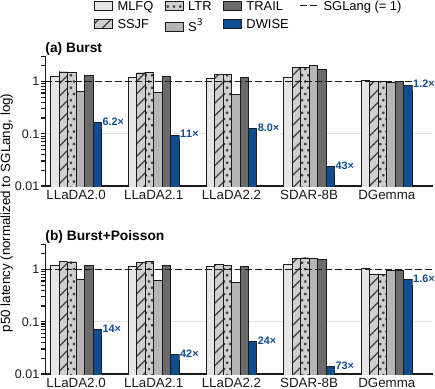}
    \caption{Normalized p50 latency relative to SGLang; lower is better, and annotations show speedup. DGemma: DiffusionGemma.}
    \label{fig:e2e-p50}
\end{figure}

\textbf{Median latency.} Figure~\ref{fig:e2e-p50} reports p50 latency. On the three LLaDA2.x models, DWISE improves p50 over SGLang by 6.2--11$\times$ under burst arrivals and by 14--42$\times$ under burst+Poisson arrivals. On SDAR-8B, the improvements are 43$\times$ and 73$\times$. These gains exceed those in mean latency because DWISE completes requests with little remaining work early, whereas the mean is dominated by a few long requests. On DiffusionGemma, p50 improves by 1.2$\times$ and 1.6$\times$. SGLang's prefill-first policy is meant to protect time to first token (TTFT), the time to a request's first output block, yet it fails even there: DWISE lowers mean TTFT on LLaDA2.1-mini by 7.8--13.0$\times$ (Appendix~\ref{app:ttft}).

\begin{figure}[t]
    \centering
    \includegraphics[width=\columnwidth]{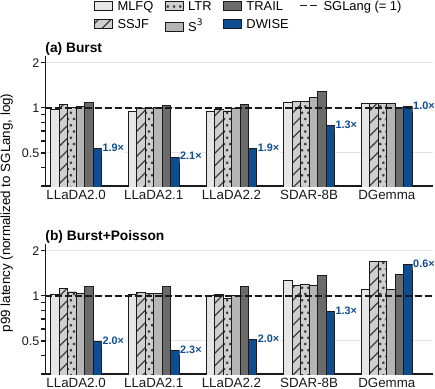}
    \caption{Normalized p99 latency relative to SGLang; lower is better, and annotations show speedup.}
    \label{fig:e2e-p99}
    \vspace{-3pt}
\end{figure}

\textbf{Tail latency.} DWISE also shortens the tail (Figure~\ref{fig:e2e-p99}), cutting p99 latency by 1.9--2.3$\times$ on the LLaDA2.x models and by 1.3$\times$ on SDAR-8B, whose throughput barely changes. SRPT serves the longest requests last, yet they still finish sooner, because mixed-phase batching keeps new prompts from stalling them and nearly doubles LLaDA2.x throughput (Table~\ref{tab:e2e}), so the backlog drains sooner. Appendix~\ref{app:category} confirms this on LLaDA2.1-mini: p99 and maximum latency drop in every category. On DiffusionGemma, where prompts are encoded in a separate pass and mixed-phase batching does not apply, DWISE can only reorder requests, so the longest ones wait longer: p99 is unchanged under burst arrivals and rises under burst+Poisson arrivals.

\begin{figure}[t]
    \centering
    \includegraphics[width=\columnwidth]{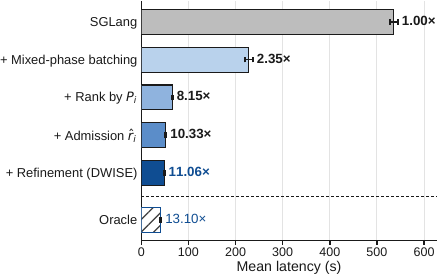}
    \caption{Cumulative speedup over SGLang as DWISE’s components are added (LLaDA2.1-mini, burst+Poisson). Rank by $P_i$: remaining prefill blocks only ($\hat{r}_i{=}0$). Oracle: ranking by true denoising rounds.}
    \label{fig:speedup-breakdown}
\end{figure}

\subsection{Ablation and Sensitivity Analysis}
\label{sec:eval-breakdown}
\label{sec:eval-ablation}

\textbf{Component breakdown.} Figure~\ref{fig:speedup-breakdown} enables the components of DWISE one at a time on LLaDA2.1-mini under burst+Poisson arrivals. Mixed-phase batching alone reduces mean latency by 2.35$\times$ over SGLang. Ranking by remaining prefill blocks alone ($\hat{r}_i{=}0$) raises the speedup to 8.15$\times$, as long-context prompts no longer run ahead of short requests. Mixed-phase batching thus creates scheduling flexibility across phases, and remaining-work ranking decides how to use it. Adding the admission-time estimate of denoising rounds raises the speedup to 10.33$\times$, and first-round refinement to 11.06$\times$ by correcting inaccurate admission estimates. An oracle ranking by true denoising rounds reaches 13.10$\times$; the predictors close 70\% of the gap to it, and DWISE reaches 84\% of its speedup.

\textbf{Round composition.} Mixed-phase batching pays off because prefill-first rounds leave most of the batch empty (Table~\ref{tab:round-composition}). On LLaDA2.1-mini, 95\% of SGLang's rounds contain only prefill blocks. Such a round usually carries a single prompt block (1.1 of 128 slots on average), so 99\% of its slots sit idle, yet these rounds take 78\% of wall-clock time. DWISE fills the idle slots with denoising blocks. It uses 113 slots per round, leaves 12\% of slots idle, and spends only 11\% of its time in prefill-only rounds. It therefore finishes the workload in 2.8K rounds instead of 64K, which is why its throughput roughly doubles in Table~\ref{tab:e2e}.

\begin{table}[ht]
\centering
\small
\caption{Serving-round composition on LLaDA2.1-mini under burst+Poisson arrivals ($B=128$). Prefill-only: rounds with only prefill blocks, as a share of all rounds and of wall-clock time.}
\label{tab:round-composition}
\setlength{\tabcolsep}{3pt}
\begin{tabular}{@{}lrcccc@{}}
\toprule
 & & \multicolumn{2}{c}{Prefill-only (\%)} & Idle & Filled \\
\cmidrule(lr){3-4}
Policy & Rounds & rounds & time & slots (\%) & slots/round \\
\midrule
SGLang & 63,926 & 94.6 & 78.2 & 96.1 & 5.0 \\
DWISE  &  2,819 &  3.0 & 11.2 & 11.6 & 113.2 \\
\bottomrule
\end{tabular}
\end{table}

\textbf{Batch former versus ranking signal.} To separate the two sources of gain, Table~\ref{tab:batch-former} runs FCFS and the learned baselines on DWISE's mixed-phase batch former, where every baseline becomes 2.1--4.1$\times$ faster than in Table~\ref{tab:e2e}. With the batch former held fixed, DWISE remains 2.5$\times$ faster than the best baseline under burst arrivals (S$^3$) and 3.3$\times$ faster under burst+Poisson arrivals (TRAIL), a gap due to the ranking alone. The batch former matters most for preemptive policies. On the prefill-first batch former, the two preemptive baselines, MLFQ and TRAIL, are slower than SGLang (0.81--0.87$\times$). Their preemption keeps up to kB = 1,024 requests resident instead of 128, yet a prefill-only round still carries only 1.6–1.7 rows, and every such round stalls all of them, so a resident request spends 5.4--5.8$\times$ more rounds running no block than under SGLang. Mixed-phase batching removes this coupling.

\begin{table}[ht]
\centering
\small
\caption{Mean latency (s) on LLaDA2.1-mini with all ranking policies on DWISE's mixed-phase batch former. FCFS is SGLang's default order. B: burst; P: burst+Poisson.}
\label{tab:batch-former}
\setlength{\tabcolsep}{4pt}
\begin{tabular}{@{}lcccccc@{}}
\toprule
 & FCFS & SSJF & S$^3$ & LTR & TRAIL & \textbf{DWISE} \\
\midrule
B & 251.7 & 282.6 & 188.4 & 271.1 & 198.5 & \textbf{76.9} \\
P & 228.0 & 256.9 & 165.1 & 248.3 & 160.7 & \textbf{48.5} \\
\bottomrule
\end{tabular}
\end{table}

\begin{figure}[t]
    \centering
    \includegraphics[width=\columnwidth]{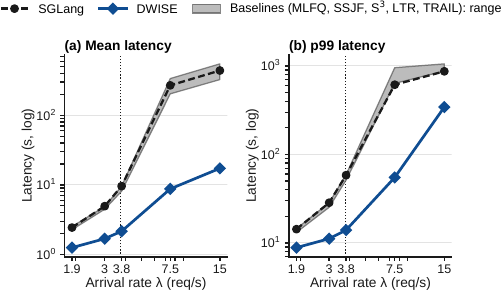}
    \caption{Latency on LLaDA2.1-mini under pure Poisson arrivals without an initial backlog. Dotted line: SGLang's throughput; grey band: the five other baselines.}
    \label{fig:load-sweep}
\end{figure}

\textbf{Load sweep.} Figure~\ref{fig:load-sweep} sweeps the arrival rate on LLaDA2.1-mini under Poisson arrivals; the five other baselines track SGLang at every rate. Below SGLang's throughput ($\lambda \le 3.8$\,req/s), DWISE reduces mean latency by 1.9--4.5$\times$ and p99 latency by 1.6--4.2$\times$. Few requests wait at this load, so the gain comes from mixed-phase batching: new prompts no longer stall denoising. Once SGLang is overloaded ($\lambda = 7.5$ and 15\,req/s), DWISE reduces mean latency by 25--31$\times$ and p99 latency by 2.5--11$\times$, because requests queue long enough for remaining-work ranking to matter.

\textbf{Refinement calibration.} Refined and admission-time estimates must share one scale, since DWISE ranks them together (Table~\ref{tab:calibration}). The diffusion-involved predictor reports the mean of a right-skewed distribution while the admission predictor regresses in log space, so its raw output ranks requests better ($\rho{=}0.890$--$0.891$) but overestimates remaining rounds: its median predicted-to-actual ratio is 1.16, against 1.03 at admission, so refined requests look longer than comparable unrefined ones. The isotonic mapping brings the ratio to 1.00 without changing $\rho$, and lowers mean latency from 80.2 to 76.9\,s under burst arrivals and from 50.9 to 48.5\,s under burst+Poisson.

\begin{table}[ht]
\centering
\small
\caption{Effect of refinement calibration on LLaDA2.1-mini, measured on the evaluation set. Ratio denotes the median predicted-to-actual remaining rounds. B: burst; P: burst+Poisson.}
\label{tab:calibration}
\setlength{\tabcolsep}{3.5pt}
\renewcommand{\arraystretch}{1.05}
\begin{tabular}{@{}llrrrr@{}}
\toprule
Arrival & Predictor & Ratio & $\rho$ & Mean (s) & p50 (s) \\
\midrule
\multirow[c]{3}{*}{B} & Admission only & 1.03 & 0.714 & 80.8 & 58.4 \\
 & Raw refinement & 1.16 & 0.890 & 80.2 & 55.2 \\
 & Calibrated & 1.00 & 0.890 & \textbf{76.9} & \textbf{51.9} \\
\midrule
\multirow[c]{3}{*}{P} & Admission only & 1.03 & 0.716 & 51.9 & 17.2 \\
 & Raw refinement & 1.16 & 0.891 & 50.9 & 15.4 \\
 & Calibrated & 1.00 & 0.891 & \textbf{48.5} & \textbf{13.7}\\
\bottomrule
\end{tabular}

\end{table}

\textbf{Refinement frequency.} Table~\ref{tab:refreq} varies how often the diffusion-involved predictor re-estimates remaining work on LLaDA2.1-mini under burst+Poisson arrivals. Refining once, after the first denoising round, gives the lowest mean and p50 latency. Refining every block, every four rounds, or every round raises mean latency to 49.5--51.0\,s. More frequent re-estimates keep reordering running requests, and preemptions per request rise from 0.34 to 1.14. Between refinements, $\hat{r}_i$ decreases by one per completed round and is clamped at zero, so an underestimated request moves to the front of the queue and runs to completion. With one refinement, 37\% of requests reach zero at least once, against at most 5.5\% otherwise. The predictor costs about 0.8\,ms of GPU time per round in every setting.

\begin{table}[t]
\centering
\small
\caption{Effect of refinement frequency on LLaDA2.1-mini under burst+Poisson arrivals. Predictor: GPU time per serving round.}
\label{tab:refreq}
\setlength{\tabcolsep}{2.5pt}
\begin{tabular}{@{}lcccc@{}}
\toprule
Refine & \textbf{Once} & Per block & Per 4 rounds & Per round \\
\midrule
Mean (s) & \textbf{48.5} & 49.5 & 51.0 & 51.0 \\
p50 (s) & \textbf{13.7} & 13.9 & 15.0 & 14.7 \\
Preemptions / request & 0.34 & 0.75 & 0.90 & 1.14 \\
Reach $\hat{r}_i=0$ (\%) & 37.0 & 5.5 & 4.9 & 0.0 \\
Predictor ($\mu$s)& 795 & 809 & 828 & 777 \\
\bottomrule
\end{tabular}

\vspace{2pt}

\caption{Generalization to unseen request categories on LLaDA2.1-mini. Within-category $\rho$ is averaged over seven categories, each held out in turn (long context excluded; see Appendix~\ref{app:loco}).}
\label{tab:loco}
\setlength{\tabcolsep}{4pt}
\begin{tabular}{@{}lcc@{}}
\toprule
 & Admission $\rho$ & Refined $\rho$ \\
\midrule
Within category, seen in training & 0.16 & 0.64 \\
Within category, held out & $-$0.06 & 0.54 \\
\midrule
All requests, MCQ seen & 0.72 & 0.89 \\
All requests, MCQ held out & 0.36 & 0.89 \\
\bottomrule
\end{tabular}
\end{table}

\textbf{Generalization to unseen categories.} Table~\ref{tab:loco} holds out one request category at a time and retrains both predictors on the rest. Prompt size features, the only input of DWISE's admission predictor, order requests within a category only weakly, even when the category is seen in training ($\rho = 0.16$ on average) and not at all once the category is held out ($\rho = -0.06$). With MCQ held out, the admission $\rho$ over all requests drops from 0.72 to 0.36. The refined estimate stays useful on held-out categories ($\rho = 0.54$) and keeps $\rho = 0.89$ over all requests, because the first-round hidden state reflects how the model denoises a request rather than what its prompt looks like. End-to-end, retraining without MCQ raises DWISE's mean latency from 47.3 to 50.9\,s in one run, a 7.6\% slowdown that still leaves DWISE 10.5$\times$ faster than SGLang (536.8\,s in the same runs).

\begin{figure}[t]
    \centering
    \includegraphics[width=\columnwidth]{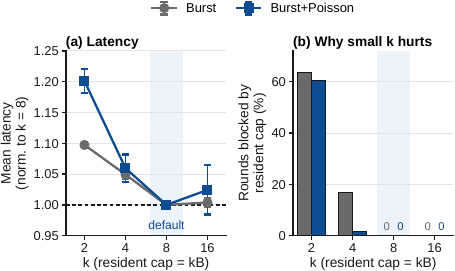}
    \caption{Effect of resident-set multiplier $k$ on LLaDA2.1-mini.}
    \label{fig:ksweep}
\vspace{-4pt}
\end{figure}

\textbf{Resident-set sensitivity.} Figure~\ref{fig:ksweep} varies the resident-set multiplier $k$, which caps KV-resident requests at $kB$. The cap trades ranking flexibility against residency: a larger resident set gives DWISE more requests to choose from in each round, but every resident request holds a KV reservation. A small cap keeps high-priority waiting requests out even when KV capacity is available. At $k=2$, the cap blocks admission in over 60\% of serving rounds and raises mean latency by about 10\% under burst arrivals and 20\% under burst+Poisson arrivals. Raising $k$ to 4 removes most of this blocking, and at $k=8$ the cap no longer blocks admission in either workload. Beyond this point, admission is governed by the KV budget rather than by the resident count, so $k=16$ brings no consistent improvement. We use $k=8$ as the default.

\begin{figure}[t]
    \centering
    \includegraphics[width=\columnwidth]{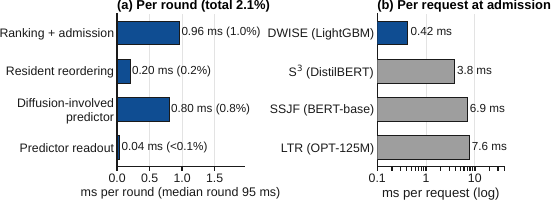}
    \caption{DWISE overhead on LLaDA2.1-mini. (a) Cost per serving round, as a share of the 95\,ms median round. (b) Admission prediction cost per request from raw prompt text (log scale).}
    \label{fig:overhead}
\end{figure}

\textbf{Overhead.} Figure~\ref{fig:overhead} breaks down DWISE's runtime cost on LLaDA2.1-mini under burst+Poisson arrivals. Figure~\ref{fig:overhead}(a) shows the cost per serving round. Ranking and admission take 0.96\,ms, reordering resident requests 0.20\,ms, and the diffusion-involved predictor 0.80\,ms, plus 0.04\,ms to read its output. Together they amount to 2.1\% of the 95\,ms median round. The predictor runs on the forward pass's stream and is read at the existing end-of-round synchronization, so it adds no synchronization point. Figure~\ref{fig:overhead}(b) shows the cost of admission-time prediction per request. DWISE's LightGBM predictor takes 0.42\,ms from raw prompt text, 9--18$\times$ less than the text encoders that S$^3$, SSJF, and LTR run on the prompt.

\section{Conclusion}
\label{sec:conclusion}

We presented DWISE, a scheduler for block-diffusion LLM serving that replaces FCFS admission and prefill priority. Its design rests on two properties that set dLLMs apart from AR models: prefill and denoising advance at the same block granularity, so neither phase needs priority over the other, and a request's remaining work becomes far more predictable after a single denoising round, whose hidden states capture model-specific information absent from the prompt. Built on these properties, DWISE lowers mean latency over SGLang's default scheduler by 5.4× on average and up to 14.9× and raises throughput by up to 2.2$\times$ across five models from three dLLM families without affecting task accuracy. Our results suggest that dLLM serving should be scheduled around how these models generate, block by block, rather than around policies inherited from AR serving.

\bibliography{example_paper}
\bibliographystyle{mlsys2025}
\clearpage

\appendix
\section{Workload Spread and Predictor Signal Across Models}
\label{app:models}

Table~\ref{tab:models} extends Figures~\ref{fig:work_signal} and~\ref{fig:work_distribution} to all five models. On every model, the first-round hidden state orders requests better than prompt features, by 0.11 to 0.16 in $\rho$. Denoising rounds span two orders of magnitude on the LLaDA2.x models and SDAR-8B. DiffusionGemma, whose single 256-token output block is denoised in ten rounds at the median, spans only 24$\times$; even there, the first-round hidden state raises $\rho$ by 0.11.

\begin{table}[h]
\caption{Denoising-round spread and predictor rank correlation across models. Rounds: denoising rounds per request over all 2,048 training requests; the minimum is 2 on every model. $\rho$: Spearman correlation with true denoising workload under five-fold cross-validation on the requests used to fit the predictors, with the estimators of Figure~\ref{fig:work_signal}: LightGBM on prompt features and a ridge probe on the mean-pooled first-round hidden state.}
\label{tab:models}
\centering
\footnotesize
\setlength{\tabcolsep}{4pt}
\begin{tabular}{lrrrrr}
\toprule
& \multicolumn{3}{c}{Rounds per request} & \multicolumn{2}{c}{$\rho$} \\
\cmidrule(lr){2-4} \cmidrule(lr){5-6}
Model & p50 & p90 & max & Prompt & 1st round \\
\midrule
LLaDA2.0-mini  & 110  & 236 & 292 & 0.657 & 0.809 \\
LLaDA2.1-mini  & 46   & 149 & 279 & 0.740 & 0.868 \\
LLaDA2.2-mini  & 53   & 189 & 296 & 0.723 & 0.863 \\
SDAR-8B        & 97.5 & 252 & 317 & 0.697 & 0.853 \\
DiffusionGemma & 10   & 18  & 48  & 0.697 & 0.809 \\
\bottomrule
\end{tabular}
\end{table}

\section{Admission Predictor Selection}
\label{app:estimator}

Table~\ref{tab:estimator-model} compares regressors for the admission-time predictor on the same three prompt features. Because the predictor runs at every admission and only needs to order requests, we compare rank correlation, measured by five-fold cross-validation on the training split, and per-request time, measured from raw prompt text on one Grace CPU thread. The small LightGBM model ranks best ($\rho = 0.740$) at the lowest cost (0.42\,ms). Larger LightGBM and XGBoost models rank no better, the MLP and ridge regression rank lower, and ExtraTrees takes 8$\times$ longer for a lower $\rho$. About half of the per-request time is tokenization; the LightGBM inference itself takes 29\,$\mu$s.

\begin{table}[h]
\centering
\small
\caption{Model selection for the admission-time workload predictor. Time is the per-request median from raw prompt text, including tokenization and feature extraction.}
\label{tab:estimator-model}
\setlength{\tabcolsep}{5pt}
\renewcommand{\arraystretch}{1.08}
\begin{tabular}{@{}lrr@{}}
\toprule
Predictor& $\rho$ & Time (ms, single request)\\
\midrule
Ridge & 0.535 & 0.42 \\
MLP (256--256) & 0.701 & 0.44 \\
XGBoost (100, depth 6) & 0.706 & 0.65 \\
LightGBM (500, 31 leaves) & 0.709 & 0.46 \\
LightGBM (1,000, 63 leaves) & 0.689 & 0.54 \\
ExtraTrees (100) & 0.722 & 3.57 \\
\textbf{LightGBM (100, 15 leaves)} & \textbf{0.740} & \textbf{0.42} \\
\bottomrule
\end{tabular}
\end{table}

\section{Seed Robustness}
\label{app:seeds}

The main results average three runs of one fixed trace per model. To check that the speedup does not depend on this trace, we generate three traces for LLaDA2.1-mini with different seeds. Each trace changes the request order and, under burst+Poisson arrivals, the Poisson arrival times. We run SGLang and DWISE once on each trace. Table~\ref{tab:seeds} reports the mean and standard deviation across the three traces. DWISE reduces mean latency by 7.10\,$\pm$\,0.14$\times$ under burst arrivals and 10.72\,$\pm$\,0.14$\times$ under burst+Poisson arrivals relative to SGLang, consistent with the single-trace results in Table~\ref{tab:e2e}.

\begin{table}[h]
\centering
\small
\caption{Seed robustness on LLaDA2.1-mini: mean $\pm$ standard deviation over three traces generated with different seeds, one run each.}
\label{tab:seeds}
\setlength{\tabcolsep}{4pt}
\begin{tabular}{@{}llrr@{}}
\toprule
Arrival & Policy & Mean latency (s) & Speedup \\
\midrule
\multirow[c]{2}{*}{B} & SGLang & 528.0\,$\pm$\,7.8 & 1.00$\times$ \\
 & \textbf{DWISE} & \textbf{74.3\,$\pm$\,0.4} & \textbf{7.10\,$\pm$\,0.14$\times$} \\
\midrule
\multirow[c]{2}{*}{P} & SGLang & 521.8\,$\pm$\,6.8 & 1.00$\times$ \\
 & \textbf{DWISE} & \textbf{48.7\,$\pm$\,1.1} & \textbf{10.72\,$\pm$\,0.14$\times$} \\
\bottomrule
\end{tabular}
\end{table}

A request's number of denoising rounds also varies slightly from run to run, because it is batched with different requests in each run. Table~\ref{tab:round-stability} compares per-request round counts across runs on LLaDA2.1-mini. Under DWISE, 98\% of requests need exactly the same number of rounds when the same trace is replayed, and 92\% across different traces; under SGLang, 37\% do across seeds. Rank correlation stays at 0.97 or above in every case, so the per-request workload that DWISE predicts is stable enough to rank requests by. The same variation is why the oracle in Figure~\ref{fig:speedup-breakdown}, which uses round counts from a profiling run, is an approximate upper bound.

\begin{table}[h]
\centering
\small
\caption{Stability of per-request denoising round counts across runs on LLaDA2.1-mini. Identical: share of requests with the same round count in both runs.}
\label{tab:round-stability}
\setlength{\tabcolsep}{4pt}
\begin{tabular}{@{}llrr@{}}
\toprule
Policy & Runs compared & Identical (\%) & $\rho$ \\
\midrule
DWISE  & same trace       & 98 & 0.998 \\
DWISE  & different traces & 92 & 0.994 \\
SGLang & different traces & 37 & 0.972 \\
\bottomrule
\end{tabular}
\end{table}

\section{Output Length Versus Remaining Work}
\label{app:fcfs}

DWISE measures work in prefill blocks and denoising rounds rather than output tokens. On LLaDA2.1-mini, the two diverge sharply. The 57\% of requests that use the full 256-token budget have the same output length but work ranging from 38 to 184 (10th to 90th percentile), so output length cannot order them. Across categories, the two even move in opposite directions: code and math requests produce more tokens than chat and instruction requests (242 vs.\ 215 on average) but need 2.6$\times$ fewer denoising rounds (45 vs.\ 115).

Output-length rankers therefore misorder the requests that carry most of the work (Table~\ref{tab:fcfs}). SSJF and LTR correlate with work overall ($\rho = 0.51$ and $0.48$) only because they rank MCQ and Hard MCQ requests first, which are 23\% of requests but 3.1\% of the work. On the other 96.9\% of the work, their scores are uncorrelated with it ($\rho = -0.02$ and $-0.07$). Because mean latency weighs a misordering by the work of the request placed first, this costs more than it saves: the mean latency of code and math requests rises from 524\,s under SGLang to 842\,s (SSJF) and 925\,s (LTR). DWISE's estimates, in contrast, stay correlated with work on these requests ($\rho = 0.54$ at admission and $0.84$ after refinement). Even its admission-time round estimate alone, without the known prefill blocks, reaches $\rho = 0.46$, so a 3-feature predictor of rounds outperforms the BERT- and OPT-based predictors of output length on these requests.

Prefill blocks can dominate the work. SDAR-8B uses 4-token blocks, so a long-context request needs 2,684 prefill blocks on average but produces only 44 output tokens. These requests are 3.7\% of requests yet 40.7\% of the work. Output-length rankers see only the short output and rank them ahead of most requests (median at the 31st--38th percentile); DWISE counts their prefill blocks and ranks them near the end (98th percentile).

\begin{table}[ht]
\centering
\small
\caption{Output-length rankers on LLaDA2.1-mini under burst arrivals ($\Delta$ averages three runs; $\rho$ uses the first). Work: prefill blocks plus denoising rounds. $\rho$: Spearman correlation between a scheduler's ranking score and work within the group; both include the exactly known prefill blocks, so $\rho$ is higher than for denoising rounds alone. ``first'': ranked ahead of nearly all other requests. DWISE adm./ref.: its score at admission and after refinement. $\Delta$: the group's contribution to the change in mean latency relative to SGLang; omitted for DWISE, whose change also includes the gain from mixed-phase batching.}
\label{tab:fcfs}
\setlength{\tabcolsep}{5pt}
\begin{tabular}{@{}lrrr@{}}
\toprule
 & (Hard) MCQ & Other & All \\
\midrule
Requests & 23\% & 77\% & 100\% \\
Work & 3.1\% & 96.9\% & 100\% \\
\midrule
$\rho$, SSJF & first & $-0.02$ & $0.51$ \\
$\rho$, LTR & first & $-0.07$ & $0.48$ \\
$\rho$, DWISE adm. & -- & $0.54$ & $0.76$ \\
$\rho$, DWISE ref. & -- & $0.84$ & $0.92$ \\
\midrule
$\Delta$ SSJF (s) & $-113$ & $+170$ & $+57$ \\
$\Delta$ LTR (s) & $-114$ & $+214$ & $+100$ \\
\bottomrule
\end{tabular}
\end{table}

\section{Multi-GPU Deployment and Output Budget}
\label{app:multigpu}

The main experiments serve each model on one GPU. To test whether DWISE carries over to multi-GPU serving, we deploy LLaDA2.1-mini with tensor parallelism over two and four GPUs (TP=2, TP=4) and with expert parallelism over four GPUs (TP=4+EP=4), where attention stays tensor-parallel and each GPU holds a quarter of the 256 experts. SGLang's dLLM path does not support pipeline parallelism. DWISE runs unchanged. Every rank runs the same scheduler on the same request state, and the diffusion-involved predictor reads the final-layer hidden state, which is replicated across ranks after the tensor-parallel all-reduce. On shorter runs, we checked that all ranks make identical scheduling decisions in every round. Both predictors are the ones trained for the single-GPU runs. All runs in Table~\ref{tab:multigpu}, including the single-GPU row, come from one batch separate from Table~\ref{tab:e2e}, with three runs per configuration.

DWISE keeps its speedup on every configuration: 10.5--12.2$\times$ under burst+Poisson arrivals and 6.7--7.1$\times$ under burst arrivals, against 11.8$\times$ and 7.4$\times$ on one GPU. For comparison, moving SGLang from one to four GPUs raises its throughput by only 1.2$\times$. Ranking and admission take 0.76--1.01\,ms and the diffusion-involved predictor 0.72--0.83\,ms per round, each about 1\% of the median round, as on one GPU. Task accuracy stays between 43.8\% and 44.5\%, and the rank correlation of refined estimates differs from the single-GPU value by at most 0.004.

\begin{table}[ht]
\caption{Mean latency (s) of LLaDA2.1-mini under multi-GPU deployment, averaged over three runs. B: burst; P: burst+Poisson; Sp.: speedup ($\times$) of DWISE over SGLang.}
\label{tab:multigpu}
\centering
\footnotesize
\setlength{\tabcolsep}{3pt}
\begin{tabular}{@{}lrrrrrr@{}}
\toprule
 & \multicolumn{3}{c}{B} & \multicolumn{3}{c}{P} \\
\cmidrule(lr){2-4}\cmidrule(lr){5-7}
Config & SGLang & DWISE & Sp. & SGLang & DWISE & Sp. \\
\midrule
1 GPU      & 543.0 & 73.0 & 7.4 & 540.2 & 45.9 & 11.8 \\
TP=2       & 498.1 & 74.5 & 6.7 & 483.5 & 46.1 & 10.5 \\
TP=4       & 457.4 & 64.7 & 7.1 & 439.4 & 36.0 & 12.2 \\
TP=4+EP=4  & 454.2 & 65.4 & 6.9 & 444.7 & 38.3 & 11.6 \\
\bottomrule
\end{tabular}
\end{table}

The main experiments also cap generation at 256 tokens. Table~\ref{tab:budget} repeats the single-GPU LLaDA2.1-mini experiment with budgets of 512, 1{,}024, and 2{,}048 tokens, one run per setting. For each budget we retrain both predictors with the unchanged pipeline on a training run at that budget; all other settings match the main experiments. Raising the budget from 256 to 2{,}048 tokens raises the mean output from 176 to 536 tokens and the mean number of denoising rounds from 62 to 180. At 2{,}048 tokens, DWISE is still 4.45$\times$ (burst) and 5.18$\times$ (burst+Poisson) faster than SGLang in mean latency and 1.74$\times$ faster at the 99th percentile, and its refined estimate keeps $\rho = 0.885$--$0.894$ at every budget. Tripling the denoising rounds raises SGLang's mean latency by only 1.3--1.5$\times$, because its time goes to prefill-only rounds, whose number is set by the prompts rather than the output length; DWISE's latency instead grows with the denoising work (2.4--3.2$\times$).

\begin{table}[ht]
\caption{Output-budget sweep on LLaDA2.1-mini. Tok.\ and Rnd.: mean output tokens and denoising rounds per request in the SGLang run. SGLang and DWISE: mean latency (s). Sp.: speedup ($\times$) of DWISE over SGLang in mean latency, with the p99 ratio in parentheses, from unrounded latencies. $\rho$: Spearman correlation between DWISE's refined estimate and the denoising rounds left after the first round. The 256-token row is rerun in the same batch. B: burst; P: burst+Poisson.}
\label{tab:budget}
\centering
\footnotesize
\setlength{\tabcolsep}{3pt}
\begin{tabular}{@{}lrrrrrrr@{}}
\toprule
 & Budget & Tok. & Rnd. & SGLang & DWISE & Sp. & $\rho$ \\
\midrule
\multirow{4}{*}{B} & 256     & 176 & 62  & 526.2 & 73.0  & 7.21 (2.12)  & 0.894 \\
                   & 512     & 295 & 100 & 607.3 & 104.9 & 5.79 (1.99)  & 0.889 \\
                   & 1{,}024 & 425 & 144 & 688.7 & 145.3 & 4.74 (1.78)  & 0.892 \\
                   & 2{,}048 & 536 & 180 & 782.1 & 175.8 & 4.45 (1.74)  & 0.886 \\
\midrule
\multirow{4}{*}{P} & 256     & 177 & 62  & 568.7 & 45.4  & 12.54 (2.45) & 0.894 \\
                   & 512     & 294 & 100 & 610.0 & 76.6  & 7.97 (2.01)  & 0.890 \\
                   & 1{,}024 & 426 & 145 & 664.1 & 113.4 & 5.86 (1.75)  & 0.891 \\
                   & 2{,}048 & 537 & 181 & 764.5 & 147.5 & 5.18 (1.74)  & 0.885 \\
\bottomrule
\end{tabular}
\end{table}

\section{Time to First Token}
\label{app:ttft}

Section~\ref{sec:motiv-unified} notes that SGLang inherits prefill-first execution from AR serving, where it protects first-token latency. For a dLLM, we measure time to first token (TTFT) as the time from a request's arrival until its first output block is committed. That block still needs its denoising rounds, so prefill-first execution delays it whenever prompts are waiting. Table~\ref{tab:ttft} reports TTFT on LLaDA2.1-mini, from the same batch of runs as the single-GPU row of Table~\ref{tab:multigpu}.

Under SGLang, mean TTFT is close to mean end-to-end latency: requests wait almost as long for their first block as for their last. DWISE lowers mean TTFT by 13.0$\times$ under burst+Poisson arrivals and 7.8$\times$ under burst arrivals, and p99 TTFT by 2.2--2.4$\times$. The median request receives its first block after 6.8\,s instead of 570.5\,s under burst+Poisson arrivals, because short requests are admitted and denoised first. The reduction holds under multi-GPU deployment: 11.6--13.7$\times$ under burst+Poisson arrivals and 7.0--7.4$\times$ under burst arrivals.

\begin{table}[h]
\caption{Time to first token (s) on LLaDA2.1-mini, from arrival until the first output block is committed, averaged over three runs. B: burst; P: burst+Poisson.}
\label{tab:ttft}
\centering
\small
\begin{tabular}{@{}llrrr@{}}
\toprule
Arrival & Policy & Mean & p50 & p99 \\
\midrule
\multirow{2}{*}{B} & SGLang & 525.2 & 541.1 & 1035.4 \\
                   & DWISE  & \textbf{67.1} & \textbf{45.2} & \textbf{472.3} \\
\midrule
\multirow{2}{*}{P} & SGLang & 521.5 & 570.5 & 983.8 \\
                   & DWISE  & \textbf{40.0} & \textbf{6.8} & \textbf{403.3} \\
\bottomrule
\end{tabular}
\end{table}

\section{Per-Category Tail Latency}
\label{app:category}

Table~\ref{tab:cat-llada} breaks down p99 and maximum latency by request category on LLaDA2.1-mini. DWISE improves both p99 and maximum latency in every category under both arrival patterns. Long-context requests bear the cost of SRPT: their prompts span over 300 prefill blocks, so DWISE runs them after shorter requests, and 44\% (burst) and 39\% (burst+Poisson) of them individually finish later than under SGLang. Their p99 and maximum latency still drop by about half, because SGLang delays them as well behind the prefill-first backlog.

\begin{table}[ht]
\centering
\caption{Per-category p99 and maximum latency (s) on LLaDA2.1-mini, SGLang $\rightarrow$ DWISE (per-run values averaged over three runs). B: burst; P: burst+Poisson.}
\label{tab:cat-llada}
\resizebox{\columnwidth}{!}{%
\footnotesize
\setlength{\tabcolsep}{2.2pt}
\begin{tabular}{@{}lcccc@{}}
\toprule
 & \multicolumn{2}{c}{B} & \multicolumn{2}{c}{P} \\
\cmidrule(lr){2-3}\cmidrule(lr){4-5}
Category & p99 & Max & p99 & Max \\
\midrule
Chat              & 1048${\to}$215 & 1050${\to}$244 & 1009${\to}$196 & 1016${\to}$224 \\
Instruction       & 1045${\to}$246 & 1049${\to}$260 & 960${\to}$219 & 962${\to}$233 \\
Code              & 1038${\to}$81 & 1041${\to}$130 & 950${\to}$44 & 952${\to}$116 \\
Math              & 1040${\to}$83 & 1043${\to}$133 & 950${\to}$52 & 954${\to}$114 \\
Multiling.\ math  & 1038${\to}$121 & 1045${\to}$125 & 953${\to}$86 & 958${\to}$105 \\
MCQ               & 1036${\to}$17 & 1039${\to}$68 & 947${\to}$10 & 948${\to}$38 \\
Hard MCQ          & 1035${\to}$6 & 1039${\to}$16 & 945${\to}$2 & 948${\to}$4 \\
Long context      & 1039${\to}$510 & 1042${\to}$511 & 954${\to}$498 & 956${\to}$498 \\
\bottomrule
\end{tabular}%
}
\end{table}

\section{Per-Category Generalization}
\label{app:loco}

Table~\ref{tab:loco-full} gives the per-category results behind Table~\ref{tab:loco}. For each of the eight request categories, we remove it from the 2,048 training requests, retrain both predictors on the remaining categories, and measure $\rho$ among that category's evaluation requests. Each cell reports $\rho$ with the category in training $\rightarrow$ held out.

The admission predictor transfers poorly. Even in distribution, prompt size features order requests within a category only weakly ($\rho = 0.04$--$0.34$). Once a category is held out, $\rho$ turns negative in five of the seven defined cases, stays near zero for chat, and is positive only for multilingual math. On long context, the held-out predictor assigns every request the same value, since these prompts are far longer than any seen in training, so $\rho$ is undefined (n/a).

The refined estimate keeps a positive $\rho$ on every held-out category. Its drop ranges from 0.02 on hard MCQ to 0.17 on long context, and the held-out $\rho$ stays between 0.40 and 0.62 except on long context. The first-round hidden state thus carries information about denoising workload that transfers across request types, whereas prompt size features do not.

\begin{table}[ht]
\centering
\small
\caption{Leave-one-category-out $\rho$ on LLaDA2.1-mini.}
\label{tab:loco-full}
\setlength{\tabcolsep}{4pt}
\begin{tabular}{@{}lcc@{}}
\toprule
Held-out category & Admission $\rho$ & Refined $\rho$ \\
\midrule
Hard MCQ & 0.200 $\rightarrow$ $-$0.154 & 0.481 $\rightarrow$ 0.463 \\
MCQ & 0.237 $\rightarrow$ $-$0.185 & 0.686 $\rightarrow$ 0.602 \\
Code & 0.251 $\rightarrow$ $-$0.202 & 0.657 $\rightarrow$ 0.539 \\
Math & 0.259 $\rightarrow$ $-$0.069 & 0.540 $\rightarrow$ 0.399 \\
Multilingual math & 0.087 $\rightarrow$ \phantom{$-$}0.163 & 0.755 $\rightarrow$ 0.623 \\
Long context & 0.338 $\rightarrow$ \phantom{$-$}n/a & 0.352 $\rightarrow$ 0.178 \\
Chat & 0.039 $\rightarrow$ \phantom{$-$}0.037 & 0.682 $\rightarrow$ 0.577 \\
Instruction & 0.053 $\rightarrow$ $-$0.033 & 0.658 $\rightarrow$ 0.561 \\
\bottomrule
\end{tabular}
\end{table}

\end{document}